# Transfer in a Connectionist Model of the Acquisition of Morphology[*]

Michael Gasser
Computer Science and Linguistics Departments,
Cognitive Science Program
Indiana University
gasser@indiana.edu

## Abstract

The morphological systems of natural languages are replete with examples of the same devices used for multiple purposes: (1) the same type of morphological process (for example, suffixation for both noun case and verb tense) and (2) identical morphemes (for example, the same suffix for English noun plural and possessive). These sorts of similarity would be expected to convey advantages on language learners in the form of transfer from one morphological category to another. Connectionist models of morphology acquisition have been faulted for their supposed inability to represent phonological similarity across morphological categories and hence to facilitate transfer. This paper describes a connectionist model of the acquisition of morphology which is shown to exhibit transfer of this type. The model treats the morphology acquisition problem as one of learning to map forms onto meanings and vice versa. As the network learns these mappings, it makes phonological generalizations which are embedded in connection weights. Since these weights are shared by different morphological categories, transfer is enabled. In a set of experiments with artificial stimuli, networks were trained first on one morphological task (e.g., tense) and then on a second (e.g., number). It is shown that in the context of suffixation, prefixation, and template rules, the second task is facilitated when the second category either makes use of the same forms or the same general process type (e.g., prefixation) as the first.

# 1 Shared Morphology and Transfer

The morphological systems of natural languages are replete with examples of the same devices used for multiple purposes. At the most abstract level, this involves the tendency for languages to rely on the same type of morphological process for diverse grammatical functions. A language that uses prefixation for noun inflection is more likely use prefixation for verb inflection than is a language that uses suffixation for noun inflection. For example, Turkish is a mainly suffixing language, while Swahili is a mainly prefixing language. Of course some languages are characterized by a variety of types of morphological processes, but even here, we may find common combinations of processes for diverse functions. Thus Semitic languages make use of template morphology[1], prefixation, and suffixation, but the combination of template and suffix is common to various syntactic categories in many of the languages.

At a more specific level, languages may make use of identical forms for different functions. This is most familiar for syncretic processes, whereby a single root undergoes the same morphological process for different functions, for example, in the possessive and regular plural forms of English nouns. But the phenomenon is more general than this; words belonging to entirely different syntactic categories may be subject to the very same processes. Thus, in English the noun plural suffix is also identical to the third person singular present tense suffix and to a number of other forms, all undergoing the same morphophonological alternation.[2] A more complex example of the common use of particular formal devices is provided by the Ethiopian Semitic language Chaha, where an elaborate palatalization process signals the second person singular feminine of the imperfect and jussive forms of verbs, an equally elaborate labialization process accompanies the third person singular masculine object suffix on verbs of all tenses and aspects, and the impersonal form of verbs in all tenses and aspects is marked by a combination of these same two processes.[3]

---

[1] A word formed by a *template* morphology process can be described as a combination of a root and a template which specifies how segments are to be intercalated between the root segments.

[2] "Morphophonological" (or "morphophonemic") "alternation" refers to the systematic phonological changes which take place in the context of particular morphological combinations. For example, the English noun plural suffix takes the form /z/ in *dogs* and /s/ in *cats* because of the voicing of the /g/ in *dog* and the lack of voicing of the /t/ in *cat*.

[3] *Palatalization* and *labialization* are processes characterized by secondary articulations accompanying phonological segments.

These similarities of form across the morphological systems of particular languages make good sense from the perspective of acquisition. If learners can get the details right for one portion of the grammar, they may not have to worry about them for another. For perception, they might develop processing strategies that focus attention for particular purposes (that is, the extraction of lexical vs. grammatical information) on the beginnings, ends, or middles of words, on recurring patterns, or on either consonants or vowels, depending on the sort of rule involved. For production, they might learn routines which are specialized for combining lexical and grammatical input in particular ways, for example, by dedicating different resources to different portions of the word being produced or by alternating consonants and vowels in particular sequences. Of course, when identical forms are involved, as opposed to more general similarities of morphological rule type (e.g., suffixation vs. prefixation), there is a trade-off with respect to comprehensibility. Faced with homophony, the listener must decide, for example, whether an English /z/ suffix signals a noun plural, a singular noun possessive, or the third person singular present tense of a verb.

I am not aware of experimental demonstrations that language learners actually do benefit from morphological similarity of the types discussed here. Still, given the comprehensibility problem, the frequency of homophony in grammatical morphology is quike striking and leads one to believe that the similarity must be conferring some advantage on learners and/or speakers. For the purposes of this paper, I will be assuming that language acquisition is facilitated by similarity in morphological rule type as well as in the actual form of the morphemes, though this facilitation still needs to be verified for human language learners. This assumption means that, given mastery of a particular form or a particular type of morphological process (e.g., suffixation) for one function, we would predict faster learning of the same form or process for some other function than without the prior learning. However, such transfer can only take place if the learner has the capacity to generalize across different morphological tasks. That is, there must be the right sort of knowledge sharing in the system to permit an advantage for the kinds of similarity we are considering.

Consider the example of English noun plural and verb present tense. To make matters simpler, let us assume that the two forms are learned in succession. Say a child has successfully learned the plural morpheme. For our purposes, this means that she has learned

1. that there are two distinct meanings, singular and plural, associated

with nouns and that these are signaled by the morphology

2. that the singular is unmarked while the plural is marked by a suffix whose precise form depends on the final segment in the stem of the noun.

Now the child is presented with the task of learning the regular present tense inflection. This requires again knowing what is signaled, in this case, distinctions of the number and person of the subject, and how it is signaled. If the learning of the latter is to be facilitated, the system must have access at this point to what it has learned about the signaling of the noun plural, as well as all other potentially relevant inflections.

In connectionist models, knowledge sharing translates into shared hardware, specifically the weighted connections that join processing units. At least some of these connections must be utilized by the two domains across which generalization is to be made. If there is complete modularity between the parts of the system dedicated to the different tasks, then no generalization is possible. In their critique of the Rumelhart and McClelland model of the acquisition of the English past tense (Rumelhart & McClelland, 1986), Pinker and Prince fault the model on these grounds, for what they call "morphological localism": the English past tense forms are learned in a network which is dedicated to this morphological task alone (Pinker & Prince, 1988). In the Rumelhart and McClelland model, as in most of the succeeding connectionist models of morphology acquisition (Daugherty & Seidenberg, 1992; MacWhinney & Leinbach, 1991; Plunkett & Marchman, 1991), morphology learning consists in learning to map a stem onto an affixed form. If such a network is trained to learn the English past tense, say, how would it benefit from this knowledge in learning the identically formed regular English past participle? Pinker and Prince explain how symbolic models distinguish morphology from phonology and how the phonological generalizations which characterize the various English -s and -ed suffixes can be captured in a small set of simple phonological rules. The problem for connectionist models, they argue, is that there is no place for phonology. As MacWhinney and Leinbach (1991) have shown, however, morphological localism is not an inherent feature of connectionist models. By adding a set of input features which distinguish different morphological categories, e.g., past tense and past participle, they give their network the capacity to generalize from one form to another.

While the authors describe only a few relevant results, MacWhinney and Leinbach's model seems to have the capacity to exhibit transfer when one suffix resembles another in some way. However, the model is clearly inadequate as a general model of morphology acquisition. The phonological input is one that presupposes an analysis of the stem into an English-specific syllabic template and the left- and right-justification of the beginnings and ends of words. Clearly something very different would be needed for a language with a radically different phonology.

More importantly, however, this model, like most of the other connectionist morphology models,[4] is based on the assumption that the learning of morphology is a matter of mapping form onto form. Certainly the ostensible task for the child is to learn to understand what is said to her and to produce words which convey what she intends, that is, to map form onto meaning and vice versa. If part of this task involves the apparently simpler task of mapping one form onto another, the child must somehow figure out what is to be mapped onto what. In any case, the child is rarely presented the two relevant forms in succession. The form-to-form mapping seems even less plausible as a component of morphology learning in the case of languages where the stem never occurs as a surface form. In learning the Japanese past tense, for example, would the child be expected to map a stem such as *nom* 'drink' onto the corresponding past form *nonda*, even though *nom* never appears in isolation in the language and is in fact not even a legitimate Japanese syllable? A stem such as Japanese *nom* constitutes an abstraction, what linguists call an "underlying representation", and if it plays a role at all in learning, it is certainly a product of the learning process rather than something to be taken as given. The situation is even more serious for non-affixing morphology, where the relevant mapping may be from an underlying sequence of consonants, e.g., Arabic *ktb* 'write', to an actual verb stem, e.g., *katab-* 'write (perfect)', which may occur only with one or another affix. A sequence such as *ktb* not only never occurs overtly; it is not even pronounceable. In sum, models which are trained to map surface linguistic forms onto other surface linguistic forms cannot constitute general models of the acquisition of morphology.[5]

In this paper, I describe a connectionist model of morphological acquisition, **Modular Connectionist Network for the Acquisition of Morphology** (MCNAM), which maps surface linguistic forms (sequences of pho-netic feature vectors which do not presuppose a language-specific phonological analysis) onto points in a lexicon/grammar space and vice versa. The model has separate modules for perception and production of words. In MCNAM, there is a place for phonology, namely, on the hidden layer of the network, and since the connections into this layer are shared by different morphological tasks, there is the potential for transfer of the type discussed above.

---

[4]The model of Cottrell and Plunkett (1991) is an exception, but it is only concerned with production, that is, with the mearning-to-form mapping. The model described here attempts to accommodate both perception and production.

[5]This is not to suggest that children cannot or do not learn mappings from one surface form to another, only that such learning is not in and of itself the learning of morphology.

The organization of the rest of the paper is as follows. First, I briefly describe the MCNAM model itself. Next, I discuss a set of experiments which investigate the capacity of the model to exhibit transfer. Finally, I consider some of the implications of the results of these experiments.

## 2 A Modular Connectionist Model of the Acquisition of Morphology

MCNAM consists of two interconnected modules, one dedicated to the perception of words, the other to their production. The basic architecture is shown in Figure 1. In the figure, boxes represent layers of processing units and arrows complete connectivity between layers. Sequential time is indicated by the overlapping boxes. Each module is a form of simple recurrent network, a network with separate input and output layers of processing units and a recurrent hidden layer of units connecting them. For production there is also a set of units which keeps an accumulated record of the network's sequence of outputs and treats this an additional input to the network. These units are not shown in the figure; instead they are indicated by the curved arrows on the SYLLABLE and PHONE layers of units. The perceptual module is trained to take a word in the form of a sequence of phones as input and to output a pattern representing the identity of the root and inflections[6] making up the word. The production module is trained to perform the reverse task. Each perception input and production output unit represents a phonetic feature, and the phones which are input to perception and output from production consist of phonetic feature vectors. On the perception output and production input layers, there are separate groups of units for the root "meaning" and for each inflectional category ("tense", "person", etc.). Representations of morphemes are either localized — each morpheme is associated with a single unit — or distributed — each morpheme is represented by a pattern of activation across a group of units.

---

[6]For simplicity's sake, I will refer to "inflectional" morphology, but what is claimed here is intended to apply to derivational morphology as well.

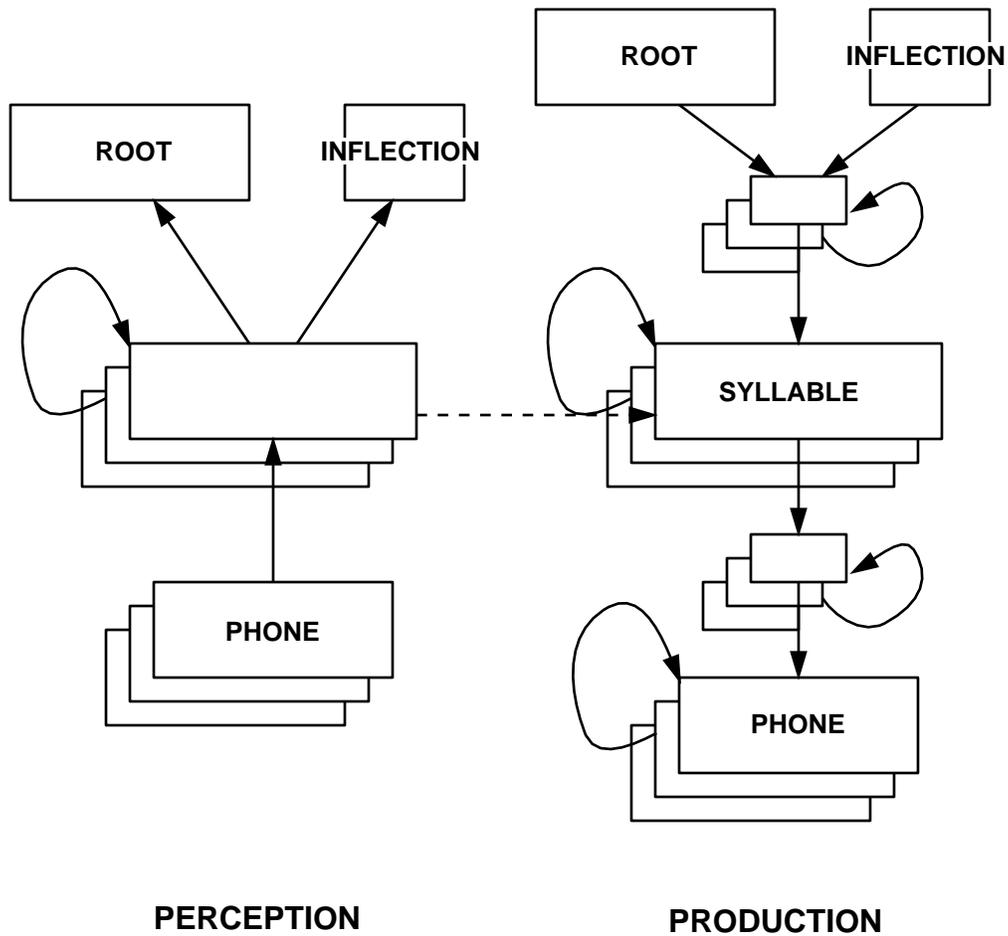

Figure 1: Architecture of MCNAM

Both word perception and word production require a short-term memory. A listener must maintain a record of what has been heard so far, and a speaker must maintain a record of what has been produced so far. In MCNAM, it is the recurrent connections which provide the short-term memory capacity. For both perception and production, the recurrent hidden-layer connections give the network access to previous hidden layer patterns, and for production, the network also has access to previous outputs.

A basic assumption behind the model is that the capacity to produce words builds on the capacity to recognize words. Word recognition is learned in the perception module in a completely supervised manner. For each input word sequence, the network is told what the correct output should be. In the perception module, phonology is learned as a side-effect as the system is trained to recognize words. Phonological knowledge takes the form of the weights on the connections from the input (phonetic) layer of processing units to the recurrent hidden layer of units. I have shown elsewhere (Gasser, 1992) that the patterns of activation appearing on this hidden layer embody generalizations about the phonological structure found in the input forms and can provide a basis for learning in the production module of the system. The link between perception and production in the current version of the model is at the level of syllables. In a trained perception network, the pattern of activation appearing on the hidden layer following the presentation of a sequence, including for example, a single syllable, constitutes a summary representation of that sequence. When input word sequences are broken into constituent syllables, the hidden-layer patterns following each syllable can be saved, yielding a sequence of distributed syllable representations. It is these syllable sequences which link the two modules of the network.

The production module is divided into two submodules, one which maps input morpheme sets (roots and inflections) onto sequences of syllables, and another which maps sequences of syllables onto sequences of phones. The former represents roughly what the child learns about how to produce words as she learns how to recognize them. The latter represents the purely phonological knowledge relating syllables to their constituent segments. The two production modules are trained separately. The syllable sequences which make up the output of one module and the input to the other are taken directly from the hidden layer of the perception module. In this paper we will only be concerned with the syllable-to-phone module within the production component of the model.

The details of network training are as follows. A **morphological task** consists of a set of words in an artificial language to be recognized or produced. For each task there is a set of roots and one or more inflectional categories, each realized as a single morpheme through the application of one regular morphological process, say, suffixation. The set of all possible combinations of roots and inflections is divided into a training set, the set of items which the network will use to adjust its weights, and a test set, the set of items which will be used to assess the network's performance but will never affect its performance directly. A training or test item is a form-meaning pair consisting on the form end of a complete word, that is, a sequence of phones, and on the "meaning" end, of the set of morphemes associated with the word form. While the meaning component of each item contains no real semantics, since it is just a list of tokens, the form-meaning association is a completely arbitrary one at the level of the individual morphemes. Also note that, like the child, the network has no direct access to the underlying representations of words.

For each training item, the perception module is presented with a sequence of input phonetic feature vectors representing the form end of the item. For supervised training, the network also requires a target. This consists of a constant pattern representing the meaning end of the training item. That is, the network is trained to recognize each of the morphemes in a word from the very beginning of the sequence. At the beginning of each word sequence, the hidden layer is re-initialized to eliminate interference from previous words. At the end of each sequence, there is a word boundary input pattern. For each input phone the hidden layer and output layer of units are activated in turn. The network's output is compared to the target pattern, an error is calculated, and the network's weights are adjusted accordingly with the familiar back-propagation learning algorithm (Rumelhart, Hinton, & Williams, 1986). For purposes of evaluating the performance of the perception module of the network, the output of the module is examined following the presentation of the word-final boundary pattern. For each morpheme, the network's response is taken to be the morpheme which its output is closest to. Performance is evaluated separately for each morphological category, that is, for the root and each inflection in a word.

I have demonstrated elsewhere (Gasser, 1994a) that the perception component has the capacity to learn prefixation, suffixation, circumfixation, infixation, deletion, mutation, and template rules.[7] I have also shown that performance is always superior with a version of the model in which root and inflection recognition are handled by separate hidden-layer modules (Gasser, 1994b). In the modular version, shown in Figure 2 the input (phone) layer is connected to both hidden-layer groups of units. However, the output root group of units is connected only to the the root hidden-layer group, and each of the output inflection groups is connected only to the inflection hidden-layer group. In this paper, all experiments make use of this modular version

---

[7]Reduplication and metathesis are not accommodated by the simple segment-based model; these would require a hierarchical version of the network which has not yet been implemented (Gasser, 1994a).

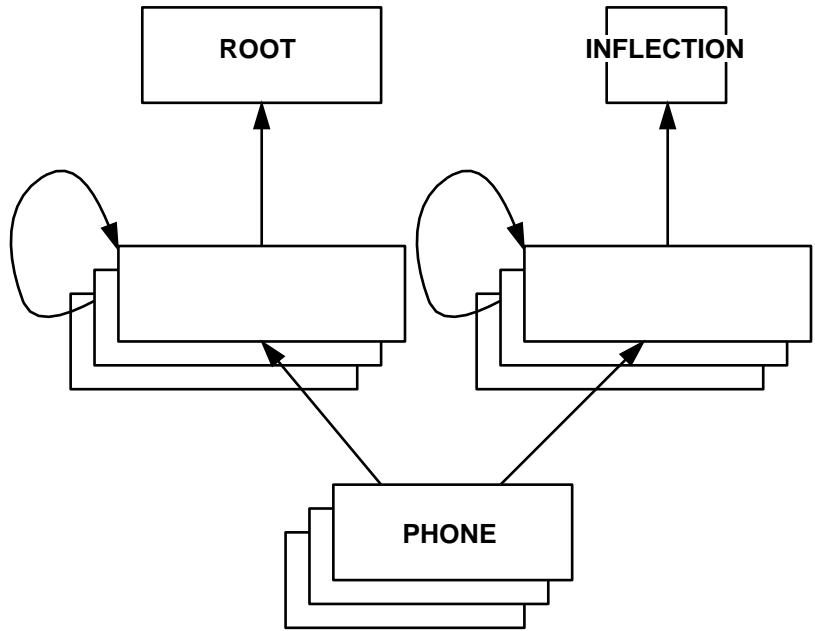

Figure 2: Modular Perception Network

of the perception component.

For production, I will describe only the training of the syllable-to-phone module. The inputs to this network are the distributed syllable representations which appear on the hidden layer of the corresponding perception network following training. For the purposes of training production, the perception network is first trained on *all* of the combinations of root and inflections, including those words which served only as test items for perception training proper. Next the syllable-to-phone production network is trained on a subset of the possible words, the remainder being set aside for testing. At the input level, each word consists of a sequence of syllables, and at the output/target level a sequence of phones. Each syllable is presented on enough time steps for the network to output the sequence of phones corresponding to that syllable. As with perception, each word is initiated with an initialized hidden-layer pattern. For production, performance is evaluated on each output of the network. The network's response is taken to be the phone which the network's output pattern is closest to.

The phonological knowledge that is embodied in the weights of both networks is also available to be used in the learning of inflectional categories other than those for which the network was originally trained. Thus there is at least the potential here for transfer from one task to another. The general question to be addressed in this paper, then, is, given a network which has been trained on a particular perception or production task, is there facilitation during training on a subsequent task?

## 3 Transfer Experiments

Consider an imaginary language learning task in which the learning of one inflectional category is followed by the learning of another, which applies either to the same syntactic category as the first or a different one. We would expect the second task to be easiest if the specific form of the inflections, as well as the type of inflectional process, were the same as for the first task. We would expect the second task to be facilitated, but less so, if the type of inflectional process, for example, prefixation, but not the specific form of the inflections, were the same. And we would expect the least facilitation, perhaps none at all, when the type of inflectional process itself differs. I describe a series of experiments which test these hypotheses for MCNAM.

In each of the experiments to be described, training and testing of the network proceeded in two **phases**. During the first phase, the network was trained on a particular task, for example, recognition of words formed with a suffixation rule. Next the trained network was presented with a second task, for example, recognition of a set of words formed with a prefixation rule. In each case, the second task required the learning of a new inflectional category. Of interest was the rate of learning of the network on this second task.

### 3.1 Perception: Prefixation, Suffixation, and Templates

The first set of experiments investigated the degree of transfer for the perception component of the network when the first and second tasks involved the same type of morphological process and further when the specific inflections were the same.

To compare performance on prefixation and suffixation, a set of 24 roots was generated, 12 of these of the form CVCVC and 12 of the form CVC. There were twelve segments in all. Prefixes and suffixes each consisted of two segments. For each inflectional category, there were 3 affixes. Two sets of prefixes (*fi-, di-, do-; be-, bu-, zi-*), and two sets of suffixes (*-if, -is, -os; -et, -ep, -up*). For example, for the root *fetos*, possible words included *fifetos*, *dofetos*, *fetosif*, and *fetosup*. In these and all other experiments reported in this paper, the set of training items consisted of 2/3 of the set of possible words, and the test set consisted of the remaining 1/3.

Pilot experiments compared performance under different conditions when the roots differed for the first and second tasks, and results were not found to differ significantly from the case where the roots were the same. Results reported here are all for a single set of roots.

For Experiments 1, there were 6 separate conditions: (a) Task 1: prefix, Task 2: suffix; (b) Task 1: suffix, Task 2: prefix; (c) Task 1: prefix (set 1), Task 2: prefix (set 2); (d) Task 1: prefix (set 1), Task 2: prefix (set 1); (e) Task 1: suffix (set 1), Task 2: suffix (set 2); (f) Task 1: suffix (set 1), Task 2: suffix (set 1).

The results are shown in Figures 3 and 4. Here, and in all succeeding plots, only the performance on the inflection recognition task is shown, and results are average performance over 10 separately trained networks. For comparison, the figures also show performance on the first task for both the prefixation and suffixation cases. Since there are always only three alternatives for the inflection recognition task, chance performance is 1/3. The results indicate clearly that perception performance on prefixation or suffixation is facilitated when the network has already been trained on the same sort of affixation and facilitated further when the affixes themselves are the same for the two tasks.

A second set of experiments examined performance of the network on words consisting of a stem and two affixes, either a prefix and a suffix or two suffixes. During the first phase, the network was trained on words containing only two morphemes, and during the second phase, the third morpheme was added. There were two conditions: (a) Task 1: prefix; Task 2: prefix and suffix; (b) Task 1: suffix; Task 2: two suffixes. The (b) condition is similar to what we might expect, for example, for the task facing a Turkish child who has learned one set of noun suffixes, say, the possessives, and is taking on another, say, the case markers.[8]

The results are shown in Figure 5. There is a clear advantage for the network learning two suffixes.

A final set of perception experiments investigated transfer for words formed with a template rule and with a suffixation rule. Since it was im-possible to use the same roots for the two kinds of rules, in all of these experiments, the set of roots for the second task differed from that for the first. For each rule type, two sets of 45 roots were generated, using an alphabet of 20 segments. For suffixation, roots took the form CVC,{footnote"C" indicates any consonant, "V" any vowel. CVCV, and CVCVC, and there were two sets of suffixes (*-if, -in, -uk; -om, -ot, -ex*). For templates, all roots consisted of three consonants, and there were two sets of templates ($C_1aC_2C_3a, C_1C_2aC_3C_3a, C_1aC_2aC_3a; C_1aC_2C_2aC_3, C_1aC_2aC_3, C_1C_2aaC_3$). Thus for the root *rng*, possible words included *ranga*, *rnagga*, and *ranaga*.

---

[8]Of course we would not expect the learning of these two categories to proceed in a strictly sequential fashion; the tasks are treated sequentially here only to allow us to separate out the effect of one task on the other.

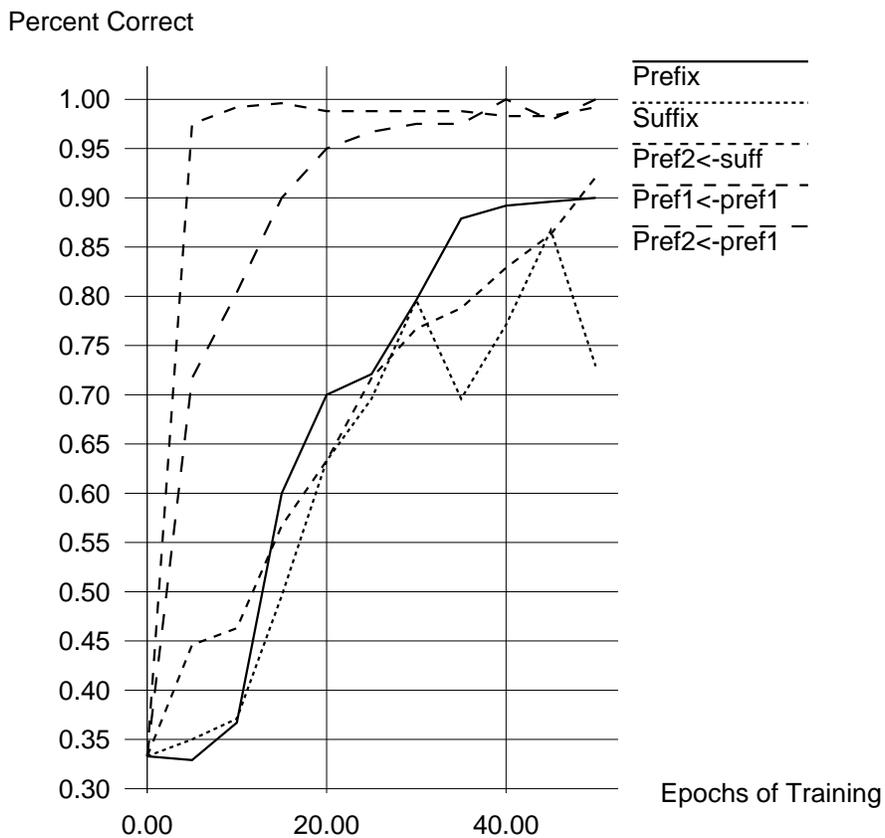

Figure 3: Experiments 1: Perception, Transfer to Prefixation

In each case, words were composed of a root and an inflection. There were four conditions: (1) Task 1: suffix, Task 2: template; (2) Task 1: template, Task 2: template; (3) Task 1: suffix, Task 2: suffix; (4) Task 1: template, Task 2: suffix.

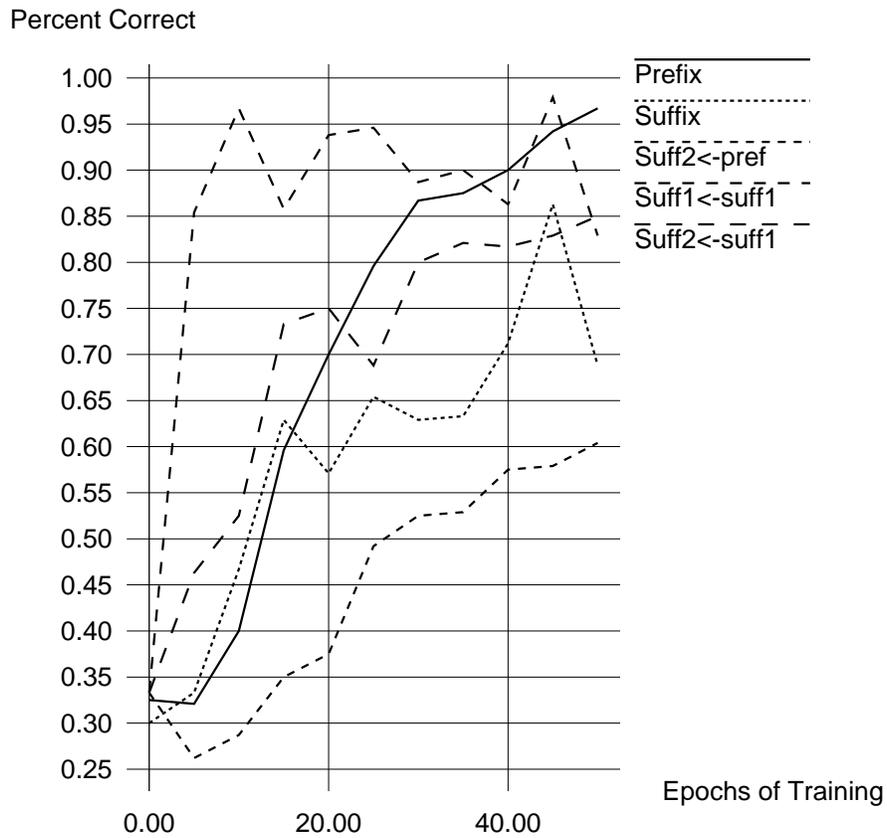

Figure 4: Experiments 1: Perception, Transfer to Suffixation

Figure 6 shows the results. Again we see a definite advantage for networks which are learning a task which is similar to one they have already learned.

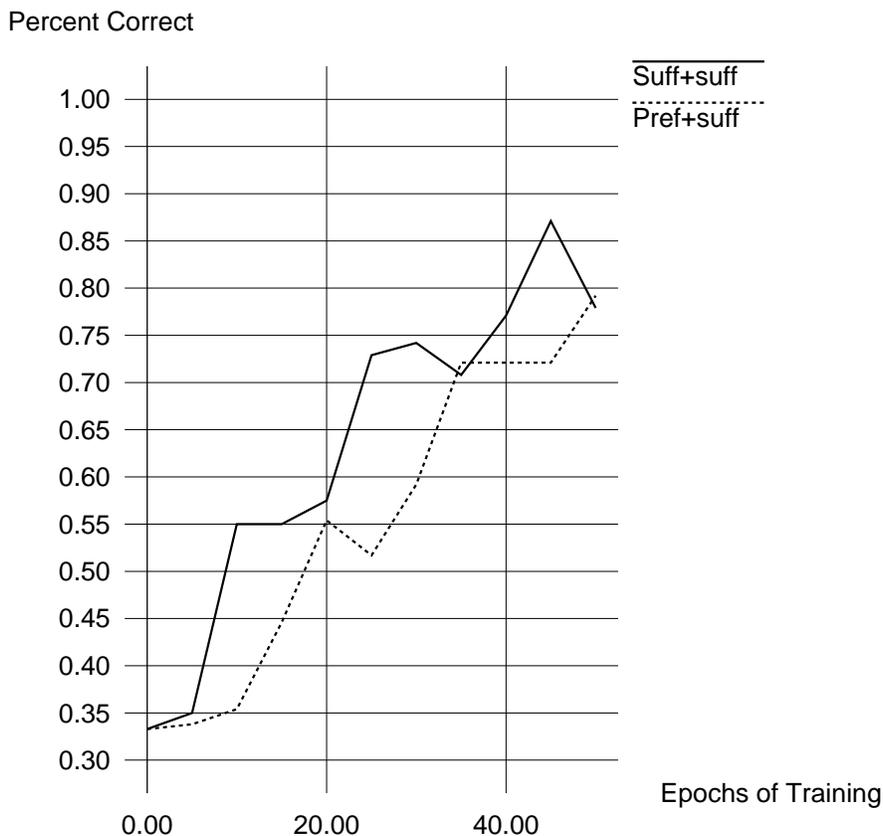

Figure 5: Experiments 2: Perception, Transfer to 2nd Affix

## 3.2 Production: Prefixation and Suffixation

For production, only the syllable-to-phone module was trained, and only prefixation and suffixation were compared. The roots and rules were identical to those used in Experiments 1. For these experiments, there were four conditions: (a) Task 1: prefix, Task 2: suffix; (b) Task 1: suffix, Task 2: prefix; (c) Task 1: prefix (set 1), Task 2: prefix (set 2); (d) Task 1: suffix (set 1), Task 2: suffix (set 2). For each condition, the corresponding perception network was first trained on all possible words. Next syllable representations were extracted from the hidden layer of the trained network; that is, the patterns of activation appearing on the hidden layer following each syllable were saved. These patterns were used as inputs to the production network, which was trained to output a sequence of phones in response to an input sequence of syllables.

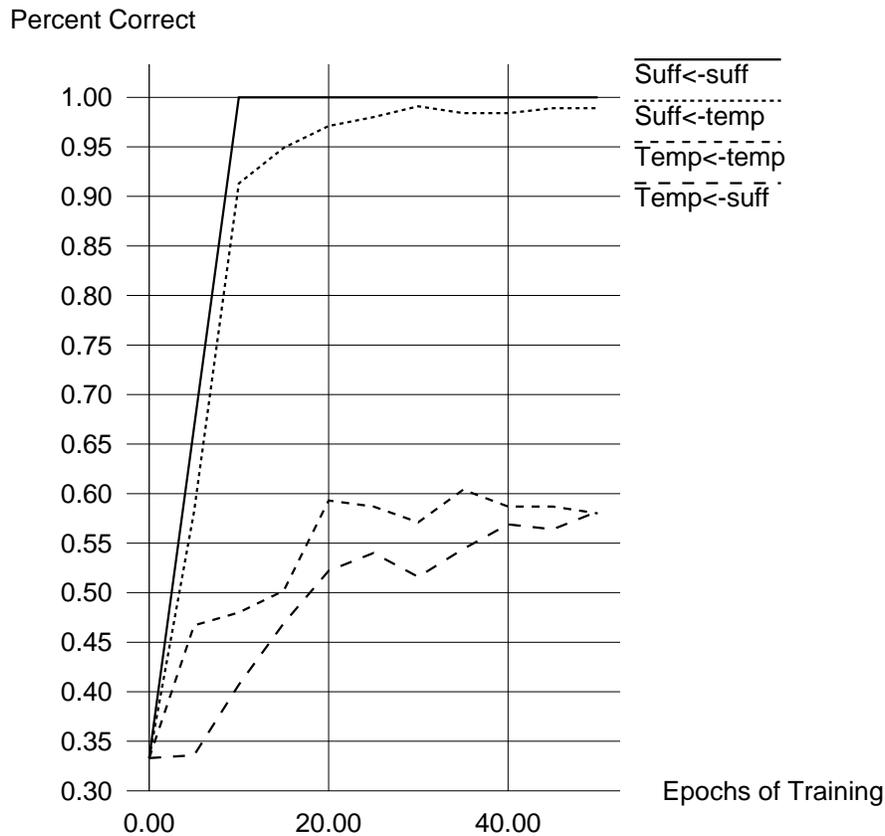

Figure 6: Experiments 3: Perception, Transfer to Suffixation and Templates

Figures 7 and 8 show results for the production experiments. Performance in each case is averaged over all of the segments. Since there are 12 phones in all, chance performance is 1/12. As with perception, we see a clear advantage for the cases in which the first and second tasks share the same type of affixation.

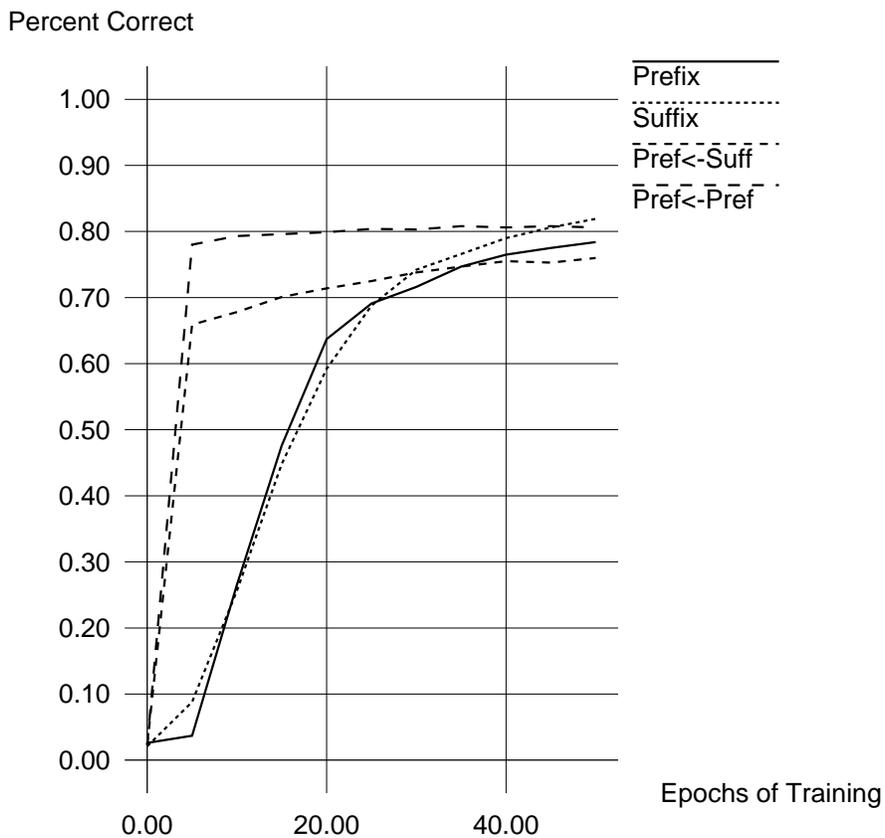

Figure 7: Experiments 4: Production, Transfer to Prefixation

## 3.3 Transfer and Vowel Harmony

As an initial investigation of the role of morphophonology in transfer for the model, two experiments examined the performance of the network trained on a suffixation rule constrained by vowel harmony.[9] Stimuli for these ex-periments were formed from a set of 42 stems (CVC and CVCVC) generated from an alphabet of 20 segments, and two separate suffixation rules. The vowels in all of the stems agreed in the feature backness. There were two separate suffixation rules, one for which the suffix vowel had to agree in backness with the stem, and for which the suffix was fixed. The two sets of suffixes constrained by harmony were *-if/-uf, -en/-on, -ik/-uk* and *-im/-um, -ex/-ox, -ep/-op*. The single set of suffixes in the fixed case (required only for the first task) was *-if, -en, -uk*. There were two conditions: (a) Task 1: harmony, Task 2: harmony; (b) Task 2: no harmony, Task 2: harmony.

---

[9] *Harmony* refers to a type of phonological constraint by which segments of a particular type within a word must agree on one or more features. In vowel harmony (as found in languages such as Turkish, for example), the vowels in a word are constrained to be similar in certain ways, for example, with respect to the height feature. Vowel harmony may interact with morphology when a language provides more than one form for a given affix so that the vowels in the affix can agree with the vowels in different sorts of roots.

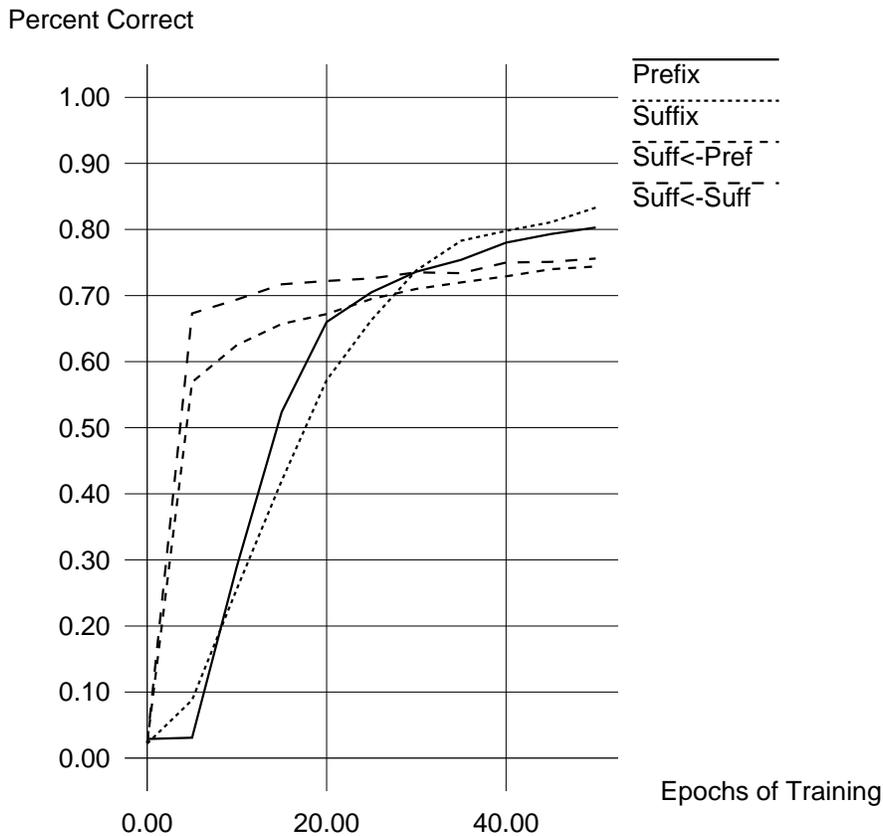

Figure 8: Experiments 4: Production, Transfer to Suffixation

Results are shown in Figure 9. There is a small, but consistent, advan-tage for the network trained initially on the harmony rule. This is the case even though the harmony rule is not inherently easier than the fixed rule.

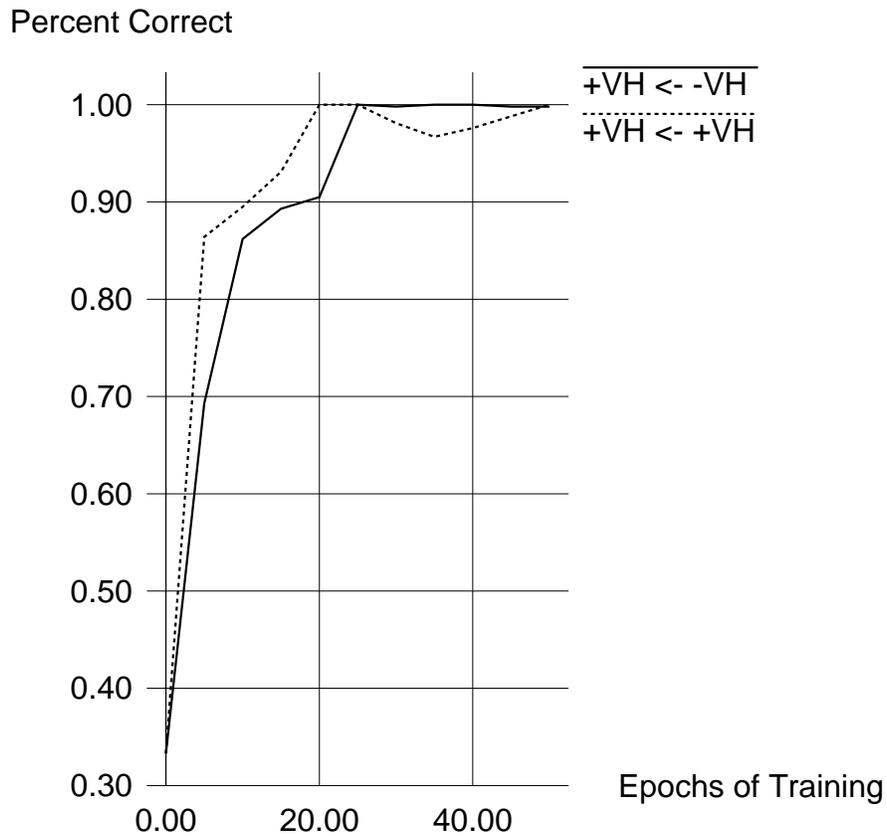

Figure 9: Experiments 5: Perception, Transfer to Suffixation with and without Vowel Harmony

## 4 Discussion and Conclusions

In summary, the experiments in the paper show that, for one particular connectionist model, performance on word recognition and production tasks is facilitated when there has been previous training on a morphologically similar task. The relevant similarity is either the general type of morpho-logical process, for example, suffixation as opposed to a template rule; or the specific form of the morphemes; or the presence of a morphophonological constraint.

In one sense this is not surprising. In the perception network (which also provides the basis for the input patterns to the production network), the hidden units, and in particular, the units in the inflection module, are involved in both of the tasks presented to the network. That is, the weights on all of the connections into these units, as well as the recurrent connections joining these units to each other, are shared by the two tasks. But a connectionist network such as this has a very large number of ways of solving a given task. It is certainly conceivable that it might make use of these resources in an idiosyncratic way, one that is no more useful for the solving of a second, superficially similar, task than any other. This is not what we find, however. As the perception network learns the first task, it finds solutions which are relatively general.

But what sorts of solutions? Unfortunately I am not yet in a position to say. One could speculate that in the case of affixation, the short-term memory built into the shared inflection weights is specialized to focus on a particular portion of a word, and preliminary evidence indicates that this is at least the case for prefixation. For templates the picture is more complicated, and further experiments are necessary to determine what aspects of template morphology the network is able to generalize about. In any case, a full-fledged connectionist theory of morphological (and phonological) learning must wait for an in-depth analysis of the network's behavior.

Much remains to be done, but these experiments provide initial evidence that connectionist models of morphological learning, which are outfitted with neither explicit roots, stems, nor affixes, are capable of generalizing from one inflectional category to another.

# References

Cottrell, G. W. & Plunkett, K. (1991). Learning the past tense in a recurrent network: acquiring the mapping from meaning to sounds. *Annual Conference of the Cognitive Science Society, 13*, 328–333.

Daugherty, K. & Seidenberg, M. (1992). Rules or connections? the past tense revisited. *Annual Conference of the Cognitive Science Society, 14*, 259–264.

Gasser, M. (1992). Learning distributed syllable representations. *Annual Conference of the Cognitive Science Society, 14*, 396–401.

Gasser, M. (1994a). Acquiring receptive morphology: a connectionist model. *Annual Meeting of the Association for Computational Linguistics, 32*, 279–286.

Gasser, M. (1994b). Modularity in a connectionist model of morphology acquisition. *Proceedings of the International Conference on Computational Linguistics, 15*, 214–220.

MacWhinney, B. & Leinbach, J. (1991). Implementations are not conceptualization: revising the verb learning model. *Cognition, 40*, 121–157.

Pinker, S. & Prince, A. (1988). On language and connectionism: analysis of a parallel distributed processing model of language acquisition. *Cognition, 28*, 73–193.

Plunkett, K. & Marchman, V. (1991). U-shaped learning and frequency effects in a multi-layered perceptron: implications for child language acquisition. *Cognition, 38*, 1–60.

Rumelhart, D. E. & McClelland, J. L. (1986). On learning the past tense of English verbs. In McClelland, J. L. & Rumelhart, D. E. (Eds.), *Parallel Distributed Processing, Volume 2*, pp. 216–271. MIT Press, Cambridge, MA.

Rumelhart, D. E., Hinton, G., & Williams, R. (1986). Learning internal representations by error propagation. In Rumelhart, D. E. & McClelland, J. L. (Eds.), *Parallel Distributed Processing, Volume 1*, pp. 318–364. MIT Press, Cambridge, MA.